\documentclass{article}
\usepackage{fullpage}
\usepackage[utf8]{inputenc}
\usepackage{multirow}
\usepackage{amsmath,amssymb,amsfonts}
\usepackage{hyperref}
\usepackage{pifont}
\usepackage{graphicx}
\usepackage{subcaption}
\usepackage{array}
\newcolumntype{P}[1]{>{\centering\arraybackslash}p{#1}}
\def\model{\theta}
\def\attackmodel{\omega}
\def\targetmodel{\Theta_c}
\def\globallossfunction{{\mathcal{L}}}
\def\locallossfunction{{\mathcal{L}_c}}
\def\lossfunction{\ell}
\def\clients{{\mathcal{C}}}
\def\dataset{{\mathcal{D}}}
\def\localdataset{\dataset_c}
\def\traindataset{\dataset_a}
\def\inputattack{\mathcal{I}}
\def\real{\mathbb{R}}

\title{Efficient passive membership inference attack in federated learning}

\author{%
  Oualid Zari \\
  Inria, Univ. C\^ote d'Azur, \\
  Sophia Antipolis, France\\
\and
  Chuan Xu \\
  Univ.  C\^ote d'Azur, Inria, CNRS, I3S \\
  Sophia Antipolis, France\\
 \and
  Giovanni Neglia \\
  Inria, Univ. C\^ote d'Azur \\
  Sophia Antipolis, France\\
    \texttt{firstname.lastname@inria.fr} \\}
  
\begin{document}

\maketitle
\begin{abstract}
In cross-device federated learning (FL) setting, clients such as mobiles cooperate with the server 
to train a global machine learning model, while maintaining their data locally.   
However, recent work shows that client's private information can still be disclosed to an adversary who just eavesdrops the messages exchanged between the client and the server. For example, the adversary can infer whether the client owns a specific data instance, which is called a passive membership inference attack~\cite{nasr}. 
In this paper, we propose a new passive inference attack that requires much less computation power and memory than existing methods. 
Our empirical results show that our attack achieves a higher accuracy on CIFAR100 dataset (more than $4$ percentage points) with three orders of magnitude less memory space and five orders of magnitude less calculations.
\end{abstract}

\section{Introduction}

    
    In recent years, it has been demonstrated that a machine learning model is vulnerable to different attacks,~e.g., membership inference attacks~\cite{shokri2017membership, truex2019demystifying}, model inversion attacks~\cite{fredrikson2015model},  attribute inference attacks~\cite{kasiviswanathan2013power}, and property inference attacks~\cite{ateniese15}, which leak sensitive information present in the training dataset. 
    The performance of these attacks depend on various factors, such as the complexity of the trained model (and then its propensity to overfit the data)
    ~\cite{truex2019demystifying} and the adversary's capabilities~\cite{shokri2017membership}, including the adversary's access to auxiliary information, like a dataset statistically similar to the client's training one~\cite{tonni2020data}.  
    
    Federated learning (FL)~\cite{mcmahan2017communication}  allows clients to participate to training without sharing their local data, but the iterative exchange of models between clients and the orchestrator can disclose additional private information.
    For instance, an adversary who has access to the mini-batch gradients computed on the client's local dataset may recover some data instances at the client~\cite{wang2019beyond, zhu2019deep, bozhao, geiping2020inverting, yin2021see}, e.g., it can reconstruct up to $97.3\%$ of a client's images to a recognizable level~\cite{yin2021see}.
    The adversary can also detect when new samples with a certain property (even unrelated to the learning task) are added  during training to the the client's local dataset~\cite{melis2019exploiting}.\footnote{Leaking this information is dangerous, for example the adversary may infer when a person starts visiting a special type of doctor.} Finally, the adversary can exploit the FL model exchanges to perform advanced \emph{client-level membership inference attacks}~\cite{nasr, zhang}.
    As a consequence, when  hospitals participate to the FL process (an increasingly popular FL use case), the adversary may infer  whether a patient has visited a particular hospital.

     In this paper, we a consider  a \emph{passive} attacker who does not interfere with the FL training process and only eavesdrops the exchanged messages. This attacker is also called \emph{honest-but-curious} and \emph{passive global attacker}  in~\cite{nasr}. 
     We show how the adversary may perform the membership inference attack  with much less  computational load (five orders of magnitude) and memory space (two/three orders of magnitude) than the state-of-the-art procedure proposed in~\cite{nasr}.     Moreover, the proposed attack can be easily adapted to the case when the auxiliary dataset only contains incomplete records (e.g., labels are missing). 

\section{Background: Passive membership inference attack for FL}\label{sec:baseline}
\paragraph{Federated learning}
In a cross-device federated learning setting, the clients (e.g.,~mobiles or IoT devices) cooperate with the server to train a global ML model $\model \in \real^d$, which minimizes the following (weighted) empirical risk over all the data owned by clients:
\begin{equation}
    \min_{\model \in \real^d} \globallossfunction(\model) = \sum_{c\in \clients} p_c\locallossfunction(\model) = \sum_{c\in \clients} p_c \left(\frac{1}{|\localdataset|} \sum_{(x,y)\in \localdataset} \lossfunction(\model, x, y)  \right),\nonumber
\end{equation}
where $\clients$ denotes the set of clients and $\localdataset$ the local dataset of client $c\in \clients$ with size $|\localdataset|$, $(x,y)\in\localdataset$ is a sample consisting of an input object $x$ and its associated label $y$, $\lossfunction(\model,x,y)$ measures the loss of the model on the sample and $p_c$ is the positive weight of client $c$, s.t.~$\sum_{c\in \clients} p_c = 1$. 
To accomplish the above learning task, many distributed learning algorithms were proposed~\cite{mcmahan2017communication, li2018federated} with FedAvg~\cite[Algo.~1]{mcmahan2017communication} being the earliest and the most popular one, which we also consider in this paper. Shortly, at each communication round $t$, the selected client $c$ receives the global model $\model^t$ from the server, updates the model following some local stochastic gradient descent updates on its dataset $\localdataset$ and sends this updated model $\model_c^t$ back to the server, who averages all the received models.

\paragraph{Adversary capabilities} 
The adversary targets a specific client $c$ and 
trains an \emph{attack model} (often a neural network) to infer whether a data point belongs to the \emph{target dataset} $\localdataset$~\cite{fredrikson2015model, truex2019demystifying, nasr}.
To this purpose, the adversary needs an auxiliary dataset $\traindataset$. 
As in~\cite{nasr}, the auxiliary dataset contains both points which do and do not belong to the target dataset (called respectively \emph{member} and 
\emph{non-member} points), i.e., 
$\traindataset \cap \localdataset \not= \emptyset$ and $\traindataset \backslash \localdataset \not= \emptyset$. 
The samples in $\traindataset \backslash \localdataset$ are generated from the same distribution of $\localdataset$.
In the FL setting, it is natural to assume that the adversary knows the architecture of the model under training, and this information is indeed needed by the attack proposed in \cite{nasr}. Our attack could instead work under a black-box model~\cite{fredrikson2015model}, where the adversary can query the targeted models with any input and receive the corresponding output (e.g., the score vector for classification problems).


\paragraph{Attack strategy for classification~\cite{nasr}}
In~\cite{nasr}, the authors consider $m$-ary classification problems.
 During training the adversary collects the models updated by $c$ at specific time instances in the set~$\mathcal T$. The collected  models $\targetmodel=\{\model_c^t, t \in \mathcal{T}\}$ are called the \emph{target models}. 


Since the adversary knows the architecture of target models $\targetmodel$, for every data sample $(x,y)\in \traindataset$, it can compute the corresponding gradients by back-propagation,~i.e., 
$\{\partial \lossfunction(\model, x,y), \forall \model\in \targetmodel \}$. Besides, it can access the loss values $\{\lossfunction(\model, x,y),  \forall\model \in \targetmodel \}$ and per-layer output values (including the last-layer output which is the prediction vector),~i.e., $\{\model^{[l]}(x)\in \real^{s(l)}, \forall l \in \{1,...,L\}, \forall \model\in \targetmodel \}$ where $L$ is the number of layers in $\model$ and $s(l)$ denotes the output size of layer $l$. These values together with a one hot encoding $\mathbf{e}_y\in \{0,1\}^m$ of the label $y$ constitute the input $\inputattack(x,y)$ to a convolutional neural network $\attackmodel$ used for membership inference prediction. 
The network is trained on $\{\inputattack(x,y), \forall (x,y) \in \traindataset\}$ by minimizing the mean square loss. 

Note that the size of $\inputattack(x,y)$ is $(d+1+\sum_{l=1}^Ls(l))\times|\mathcal T|+m$, which is extremely large for deep neural networks. For instance, when training the ResNet-110 in~\cite{nasr}, even when the authors consider only the last four layers' gradients and last three layers' outputs in $\inputattack(x,y)$, there are still more than 1.6 million parameters for each target model. Correspondingly, the neural network $\attackmodel$ needs to be large as well, e.g., the original implementation requires 256~MB.\footnote{The open source code is available at \url{https://github.com/SPIN-UMass/MembershipWhiteboxAttacks/blob/master/ATTACK-ALEXNET-grad_fed_local.py}.} In addition, the adversary incurs a high computational load to compute the $|\mathcal T|\times |\traindataset|$ gradients in $\{\inputattack(x,y), \forall (x,y)\in \traindataset \}$, as  gradients' back-propagation computation is much slower than the forward pass for deep neural networks. For instance, to achieve the attack on FL training of  AlexNet on CIFAR100, the adversary computes $5\times 20000$ gradients~\cite[Table XI and XII]{nasr}, which takes at least 1.5 hours on a NVIDIA GeForce GTX 1050 Ti.~\footnote{We used the optimized per-sample gradient calculation package offered by Opacus~\cite{opacus}.}

\section{Efficient passive membership inference attack}
\begin{figure}
    \centering
    \includegraphics[width=\textwidth]{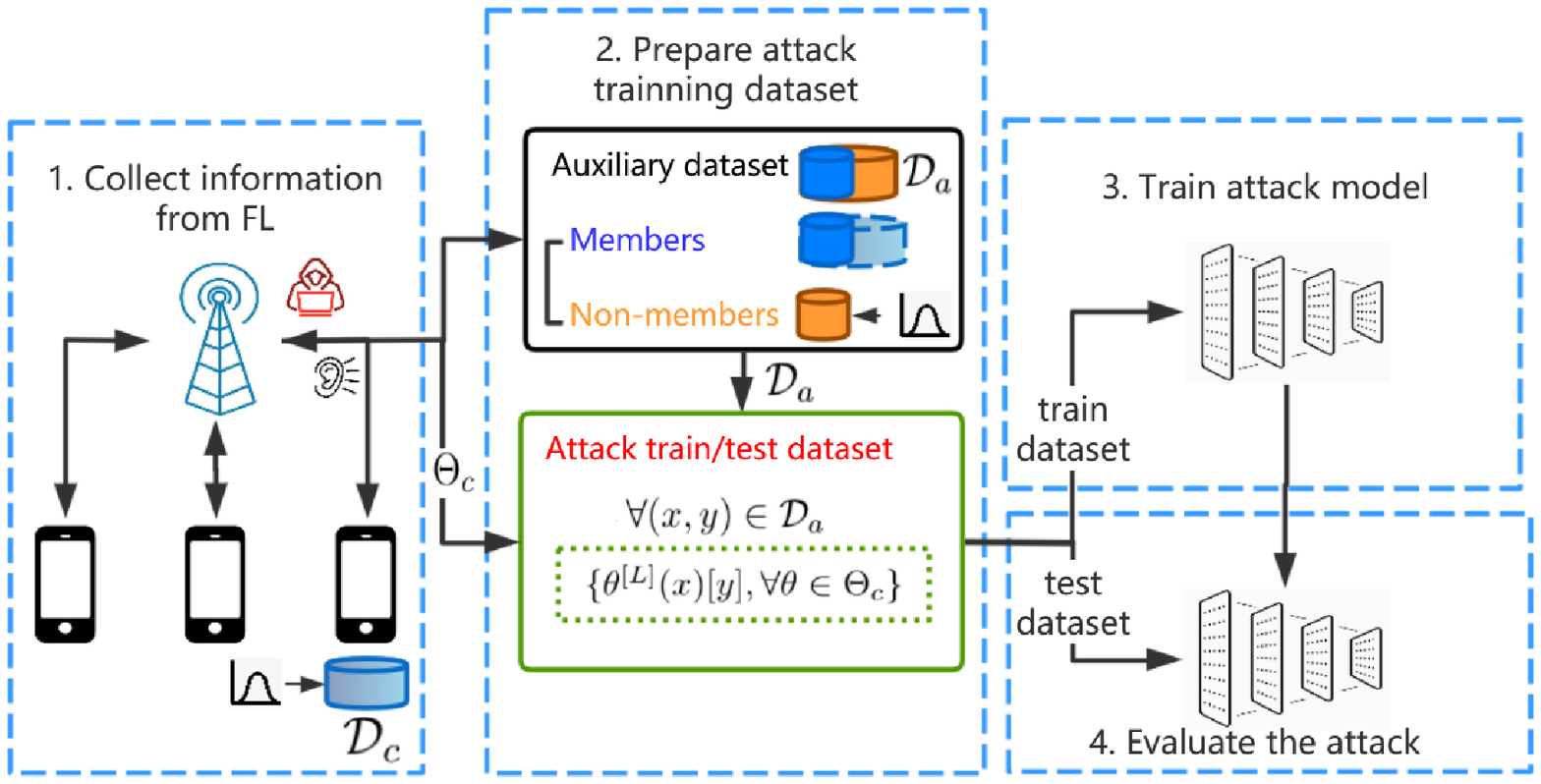}
    \caption{The procedure of our membership inference attack on federated learning. $\targetmodel$ are the set of target models, $\localdataset$ is the target dataset, $\traindataset$ is the auxiliary dataset and $\model^{L}(x)[y]$ denotes the score of instance $x$ for label $y$.}
    \label{fig:attack}
\end{figure}

In this section, we propose an efficient passive inference attack for FL which releases the adversary from the high computational burden and large memory space requirement.  Our attack is depicted in Figure~\ref{fig:attack}. 

If the adversary attacks client $c$, it starts by collecting the target models $\targetmodel = \{\model_c^t, t\in \mathcal T \}$ exchanged between client $c$ and the server. Then, for every data sample $(x,y)\in \traindataset$, it computes the score assigned by target models to the correct label $y$ for the input $x$. The input of the attack model only includes $ \{\model^{[L]}(x)[y], \forall \model \in \targetmodel \}$, where $\alpha[i]$ indicates the $i^{\textrm{th}}$ element in the vector $\alpha$. This choice is motivated by the empirical observation that the temporal evolutions of true label scores for members and non-members data points (i.e., those in ) are easily distinguishable (see Fig.~\ref{fig:distribution_prediction} in Appendix). 
Since target models corresponds to different time instants in the training process, the input can be a time series. 
We choose then as attack model a fully convolutional network, which is a suited network architecture for classifying time series~\cite{Wang2017TimeSC}.
The attack model's architecture is shown in Fig.~\ref{fig:attack_model_arch} in Appendix and it is trained minimizing the usual cross-entropy loss.

In comparison to the state-of-the-art attack in~\cite{nasr}, our approach requires a much smaller input of size~$|\mathcal T|$, independently of the size of the target model. As a result, while experiments in~\cite{nasr} are limited to consider $5$ target models, because of memory constraints, our attack has not such limit and can take into account a finer-grained temporal evolution. Also the attack model architecture in~\cite{nasr} considers the input as a flat vector, while our architecture is designed to explicitly capture the FL training dynamics, which may expose more information about data point's membership.

As most of the related work, we have assumed that samples in the auxiliary dataset $\traindataset$ consist of input-label pairs, but labels may contain  particularly sensitive information and then be better protected.
Our attack can be easily adapted to deal with the case when the adversary has no access to labels.
Indeed, it is sufficient to replace the attack input $\model^{[L]}(x)[y]$ 
with the entropy of the score  vector $\model^{[L]}(x)$ or with its maximum value. 
The size of the attack model's input does not change, but one can expect the attack accuracy to decrease as less information is available to the adversary. This is confirmed by our experiments in  Table~\ref{tab: without_konwledge_y} in the appendix.

\section{Experiments}
We evaluate our attacks on two datasets: one is CIFAR100 which contains 60,000 images for 100 different classes; the other one is Purchase100 which contains 197,324 shopping records for 600 products with customers clustered into 100 classes on the basis of the similarity of their purchases. 

For a fair comparison with the baseline in~\cite{nasr}, we consider the same FL scenario with 
4 clients  and data distributed uniformly at random among the clients. The observed epochs are $\mathcal T=\{100, 150, 200, 250, 300\}$ for the CIFAR100 dataset, and $\mathcal T=\{40, 60, 80, 90, 100\}$ for the Purchase100 dataset.\footnote{
    The number of communication rounds and training epochs coincide, as each client processes one local epoch at each communication round.
}
More details on our experimental setup are provided in the appendix.

The attack performance is evaluated in terms of accuracy of membership inference on a test auxiliary dataset.
Table~\ref{tab:compare_baseline} shows that our attack requires 2 to 3 orders of magnitued less memory space (the attack model's size) and at least 5 orders of magnitude less computation (Multiply–Accumulate Operations, MACs). For CIFAR100, the accuracy of our attack is at least $5\%$ higher than the baseline. Although our attack is less accurate on Purchase100, we can take advantage of the smaller memory footprint to increase the number of epochs considered $|\mathcal T|$. For example, for $|\mathcal T|=30$, the accuracy of our attack increases to $62.3\%$ with the same memory requirement ($1.06$~MB) and $31$~KMACs, still more than 20,000 times less operations than the baseline.  

Figure~\ref{fig:accuracy_vs_epoch} shows the the membership inference attack accuracy decreases when labels are not available, but it is still larger than $75\%$ when the adversary trains the model over the latest $10$ epochs.
The figure also illustrates client model's train and test accuracy over time and suggests that, in the setting considered in~\cite{nasr}, the model is overfitting the dataset and the FL orchestrator may have stopped the training earlier. Intuitively, overfitting can help the adversary as the model memorizes the training samples. Table~\ref{tab:compare_baseline_epoch} confirms this intuition as the accuracy of both attacks decreases when training stops earlier. When overfitting is prevented, our attack is even better than the baseline with up to $11.3\%$ accuracy increase ($6.6$ percentage points) (see Table~\ref{tab:compare_baseline_epoch}).


\begin{table}[t]
\centering
\setlength\tabcolsep{1.5pt}
    \small
    \begin{tabular}{|c|c|*{8}{P{12mm}|}}
    \hline
         \multirow{2}{*}{Dataset}&  \multirow{2}{*}{Model type}& \multicolumn{2}{c|}{$\model^T$ accuracy}&  \multicolumn{2}{c|}{Attack accuracy} &  \multicolumn{2}{c|}{Memory (MB)}&  \multicolumn{2}{c|}{MACs}\\ 
        \cline{3-10}
         & & Train & Test & \textbf{Ours}&Baseline& \textbf{Ours}& Baseline& \textbf{Ours}& Baseline\\
         \hline 
         \multirow{2}{*}{CIFAR100}& AlexNet& 99\% &36\% &89.5\% & 85.1\% & 1.06 & 1053  & 6.66K & 1.44G\\
         & DenseNet & 100\% & 55\% & 84.2\% & 79.2\% & 1.06 & 1405 & 6.66K & 1.93G \\
         \hline
         Purchase100& Fully connected& 93\% &  82\% &  60.1\% &  72.4\% & 1.06 & 527 & 6.66K & 0.72G\\
         \hline
    \end{tabular}
    \caption{Performance comparison of our passive membership inference attack with the baseline~\cite{nasr}. FL training with 4 clients  spanning $300$ epochs for different datasets and model architectures. Attack accuracy is averaged over all the clients.}
    \label{tab:compare_baseline}
\end{table}

\begin{figure}[t]
    \centering
    \begin{minipage}{0.45\textwidth}
    \setlength\tabcolsep{1.5pt}
    \captionsetup{type=table}
    \centering
        \begin{tabular}{|c|c|c|}
        \hline
        \multirow{2}{*}{\textbf{Observed Epochs}} & \multicolumn{2}{c|}{\textbf{Attack Accuracy}} \\ \cline{2-3} 
                                & Ours   & Baseline \\ \hline
        5, 10, 15, 20, 25       & 64.0\% & 57.4\%   \\
        10, 20, 30, 40, 50      & 82.2\% & 76.5\%   \\
        50, 100, 150, 200, 250  & 86.4\% & 79.5\%   \\
        100, 150, 200, 250, 300 & 89.5\% & 85.1\%   \\ \hline
        \end{tabular}
        \caption{Effect of the adversary's observed epochs on attack accuracy. Four clients trains  AlexNet to classify CIFAR100 dataset.}
        \label{tab:compare_baseline_epoch}
    \end{minipage}
    \begin{minipage}{0.5\textwidth}
    \centering
    \includegraphics[width=8cm]{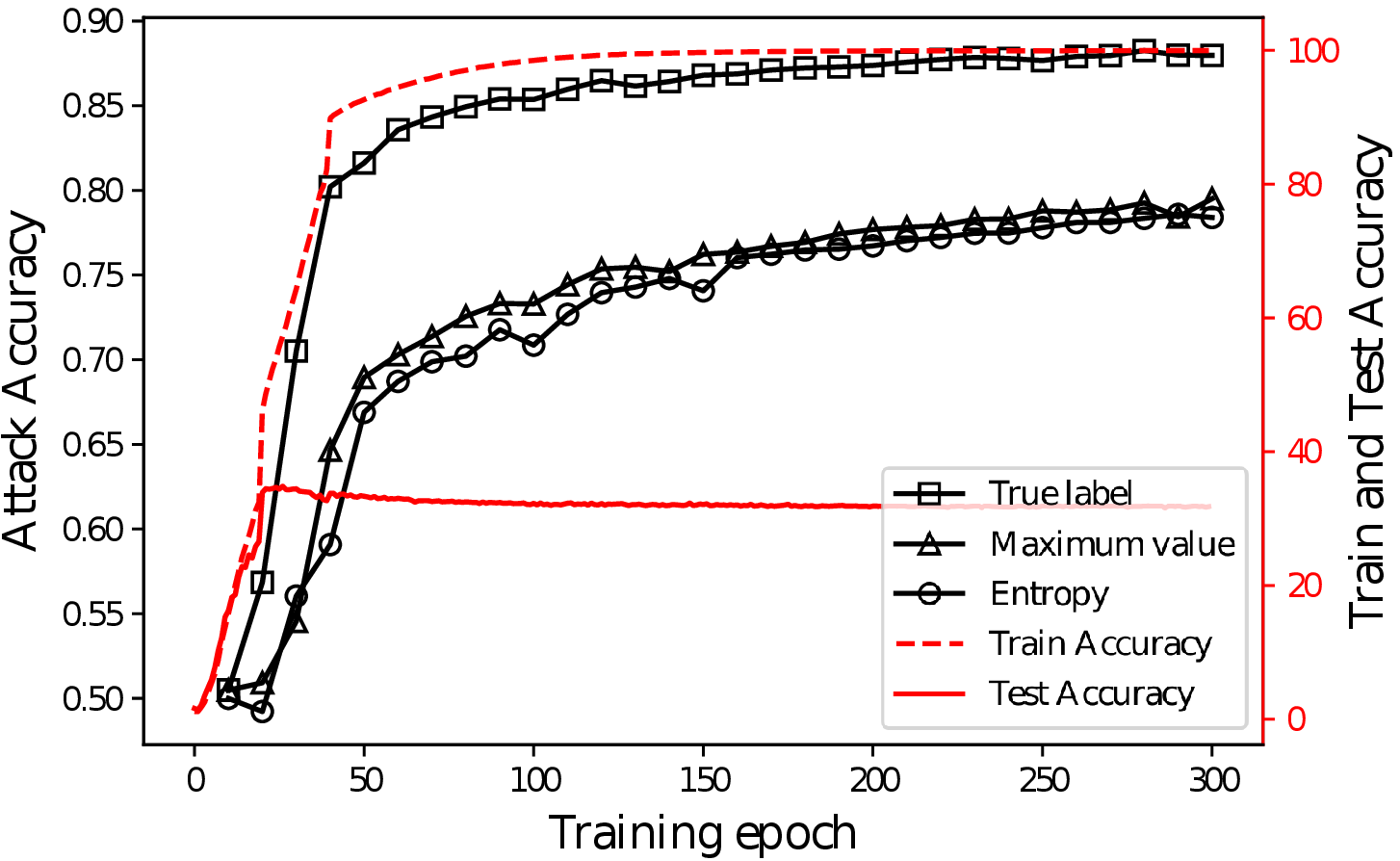}
    \centering
    \caption{Attack accuracy and train/test client's model accuracy over time $t$.
    Four clients train  AlexNet to classify CIFAR100 dataset. The attack model is trained with $\mathcal T= \{t-9, t-8,\ldots, t\}$.
    }
    \label{fig:accuracy_vs_epoch}
    \end{minipage}
\end{figure}

\bibliographystyle{plain}
\bibliography{ref.bib}

\begin{thebibliography}{10}

\bibitem{opacus}
Opacus, https://github.com/pytorch/opacus.

\bibitem{ateniese15}
Giuseppe Ateniese, Luigi~V. Mancini, Angelo Spognardi, Antonio Villani,
  Domenico Vitali, and Giovanni Felici.
\newblock Hacking smart machines with smarter ones: How to extract meaningful
  data from machine learning classifiers.
\newblock {\em Int. J. Secur. Netw.}, 10(3):137–150, September 2015.

\bibitem{fredrikson2015model}
Matt Fredrikson, Somesh Jha, and Thomas Ristenpart.
\newblock Model inversion attacks that exploit confidence information and basic
  countermeasures.
\newblock In {\em Proceedings of the 22nd ACM SIGSAC conference on computer and
  communications security}, pages 1322--1333, 2015.

\bibitem{geiping2020inverting}
Jonas Geiping, Hartmut Bauermeister, Hannah Dr{\"o}ge, and Michael Moeller.
\newblock Inverting gradients--how easy is it to break privacy in federated
  learning?
\newblock {\em NIPS}, 2020.

\bibitem{kasiviswanathan2013power}
Shiva~Prasad Kasiviswanathan, Mark Rudelson, and Adam Smith.
\newblock The power of linear reconstruction attacks.
\newblock In {\em Proceedings of the twenty-fourth annual ACM-SIAM symposium on
  Discrete algorithms}, pages 1415--1433. SIAM, 2013.

\bibitem{li2018federated}
Tian Li, Anit~Kumar Sahu, Manzil Zaheer, Maziar Sanjabi, Ameet Talwalkar, and
  Virginia Smith.
\newblock Federated optimization in heterogeneous networks.
\newblock {\em MLSys}, 2020.

\bibitem{mcmahan2017communication}
Brendan McMahan, Eider Moore, Daniel Ramage, Seth Hampson, and Blaise~Aguera
  y~Arcas.
\newblock Communication-efficient learning of deep networks from decentralized
  data.
\newblock In {\em Artificial Intelligence and Statistics}, pages 1273--1282.
  PMLR, 2017.

\bibitem{melis2019exploiting}
Luca Melis, Congzheng Song, Emiliano De~Cristofaro, and Vitaly Shmatikov.
\newblock Exploiting unintended feature leakage in collaborative learning.
\newblock In {\em 2019 IEEE Symposium on Security and Privacy (SP)}, pages
  691--706. IEEE, 2019.

\bibitem{nasr}
Milad Nasr, Reza Shokri, and Amir Houmansadr.
\newblock Comprehensive privacy analysis of deep learning: Passive and active
  white-box inference attacks against centralized and federated learning.
\newblock In {\em 2019 IEEE Symposium on Security and Privacy (SP)}, pages
  739--753, 2019.

\bibitem{shokri2017membership}
Reza Shokri, Marco Stronati, Congzheng Song, and Vitaly Shmatikov.
\newblock Membership inference attacks against machine learning models.
\newblock In {\em 2017 IEEE Symposium on Security and Privacy (SP)}, pages
  3--18. IEEE, 2017.

\bibitem{tonni2020data}
Shakila~Mahjabin Tonni, Dinusha Vatsalan, Farhad Farokhi, Dali Kaafar, Zhigang
  Lu, and Gioacchino Tangari.
\newblock Data and model dependencies of membership inference attack.
\newblock {\em arXiv preprint arXiv:2002.06856}, 2020.

\bibitem{truex2019demystifying}
Stacey Truex, Ling Liu, Mehmet~Emre Gursoy, Lei Yu, and Wenqi Wei.
\newblock Demystifying membership inference attacks in machine learning as a
  service.
\newblock {\em IEEE Transactions on Services Computing}, 2019.

\bibitem{wang2019beyond}
Zhibo Wang, Mengkai Song, Zhifei Zhang, Yang Song, Qian Wang, and Hairong Qi.
\newblock Beyond inferring class representatives: User-level privacy leakage
  from federated learning.
\newblock In {\em IEEE INFOCOM 2019-IEEE Conference on Computer
  Communications}, pages 2512--2520. IEEE, 2019.

\bibitem{Wang2017TimeSC}
Zhiguang Wang, Weizhong Yan, and T.~Oates.
\newblock Time series classification from scratch with deep neural networks: A
  strong baseline.
\newblock {\em 2017 International Joint Conference on Neural Networks (IJCNN)},
  pages 1578--1585, 2017.

\bibitem{yin2021see}
Hongxu Yin, Arun Mallya, Arash Vahdat, Jose~M Alvarez, Jan Kautz, and Pavlo
  Molchanov.
\newblock See through gradients: Image batch recovery via gradinversion.
\newblock In {\em Proceedings of the IEEE/CVF Conference on Computer Vision and
  Pattern Recognition}, pages 16337--16346, 2021.

\bibitem{zhang}
Jingwen Zhang, Jiale Zhang, Junjun Chen, and Shui Yu.
\newblock Gan enhanced membership inference: A passive local attack in
  federated learning.
\newblock In {\em ICC 2020 - 2020 IEEE International Conference on
  Communications (ICC)}, pages 1--6, 2020.

\bibitem{bozhao}
Bo~Zhao, Konda~Reddy Mopuri, and Hakan Bilen.
\newblock idlg: Improved deep leakage from gradients.
\newblock {\em CoRR}, abs/2001.02610, 2020.

\bibitem{zhu2019deep}
Ligeng Zhu, Zhijian Liu, and Song Han.
\newblock Deep leakage from gradients.
\newblock In {\em Advances in Neural Information Processing Systems}, pages
  14774--14784, 2019.

\end{thebibliography}
\newpage

\section*{Appendix}

\subsection*{Figures for Section 3}

    \begin{figure}[ht]
    \includegraphics[width=8cm]{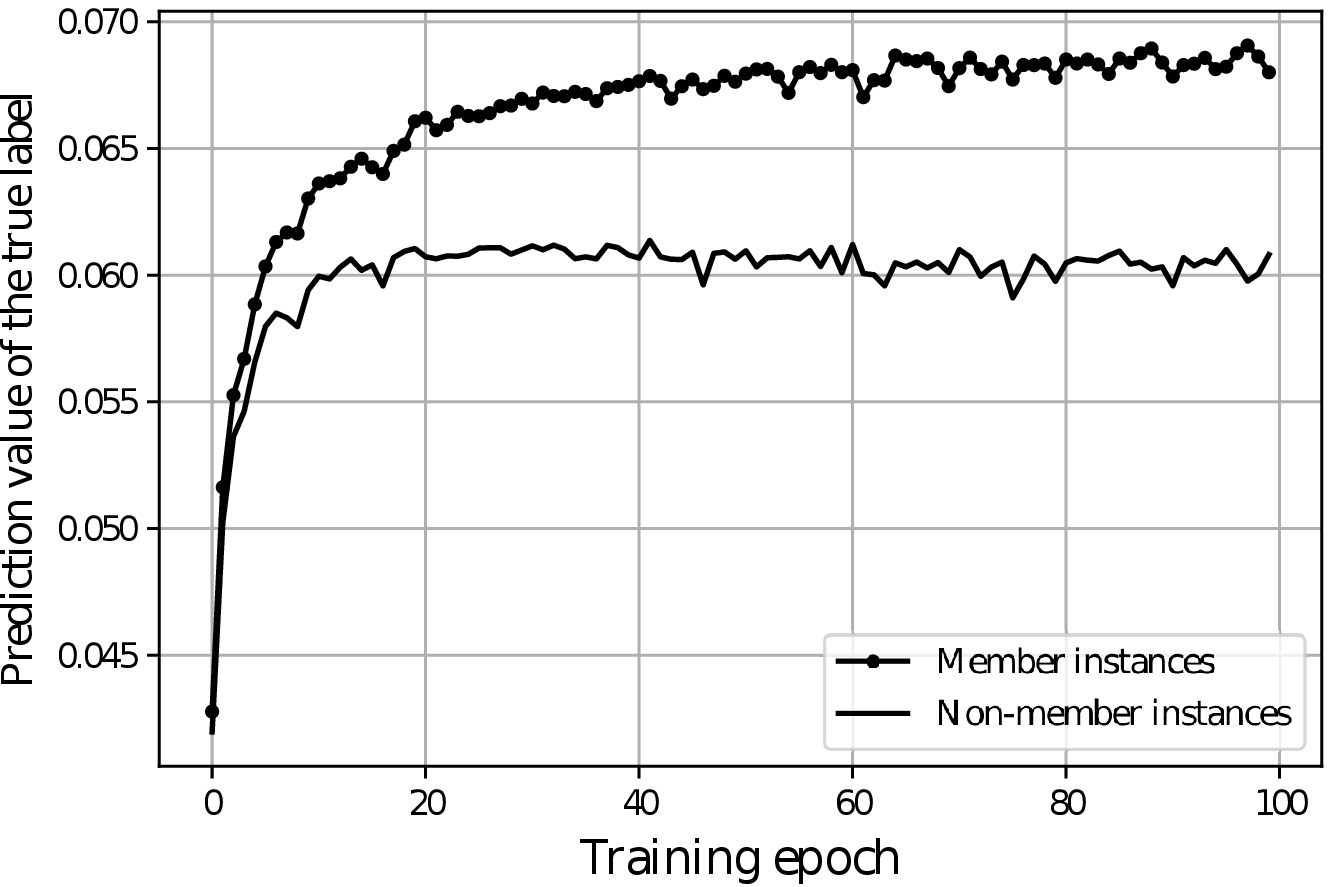}
    \centering
    \caption{Average prediction value of the true label of  Purchase100 dataset for members and non-members. }
    \label{fig:distribution_prediction}
    \end{figure}

\begin{figure}[h!]
  \centering
  \begin{subfigure}[b]{0.4\columnwidth}
    \centering
    \includegraphics[width=\linewidth]{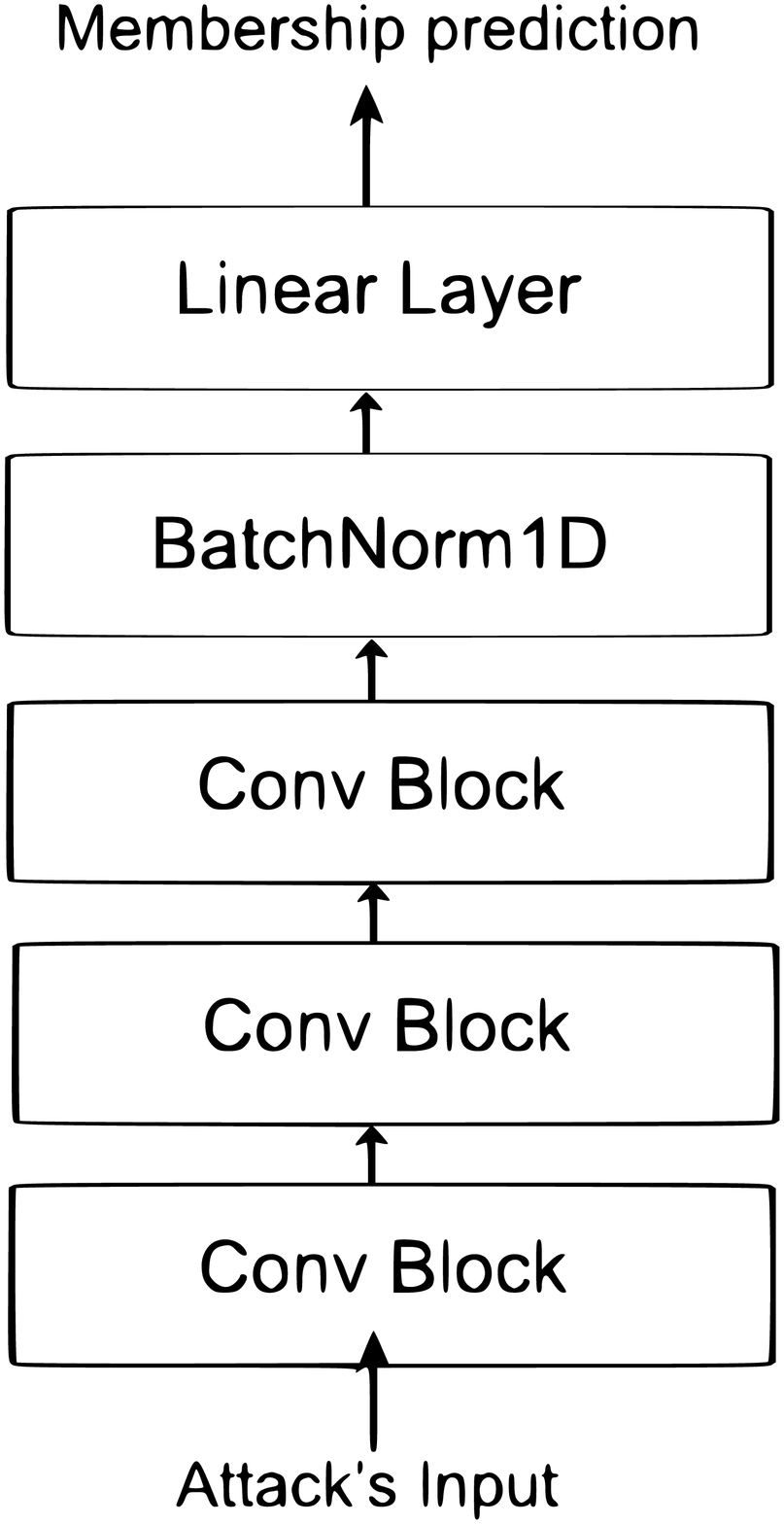}
    \caption{model's architecture}
    \label{fig:archiAM}
  \end{subfigure}%
  \begin{subfigure}[b]{0.4\columnwidth}
    \centering
    \includegraphics[width=\linewidth]{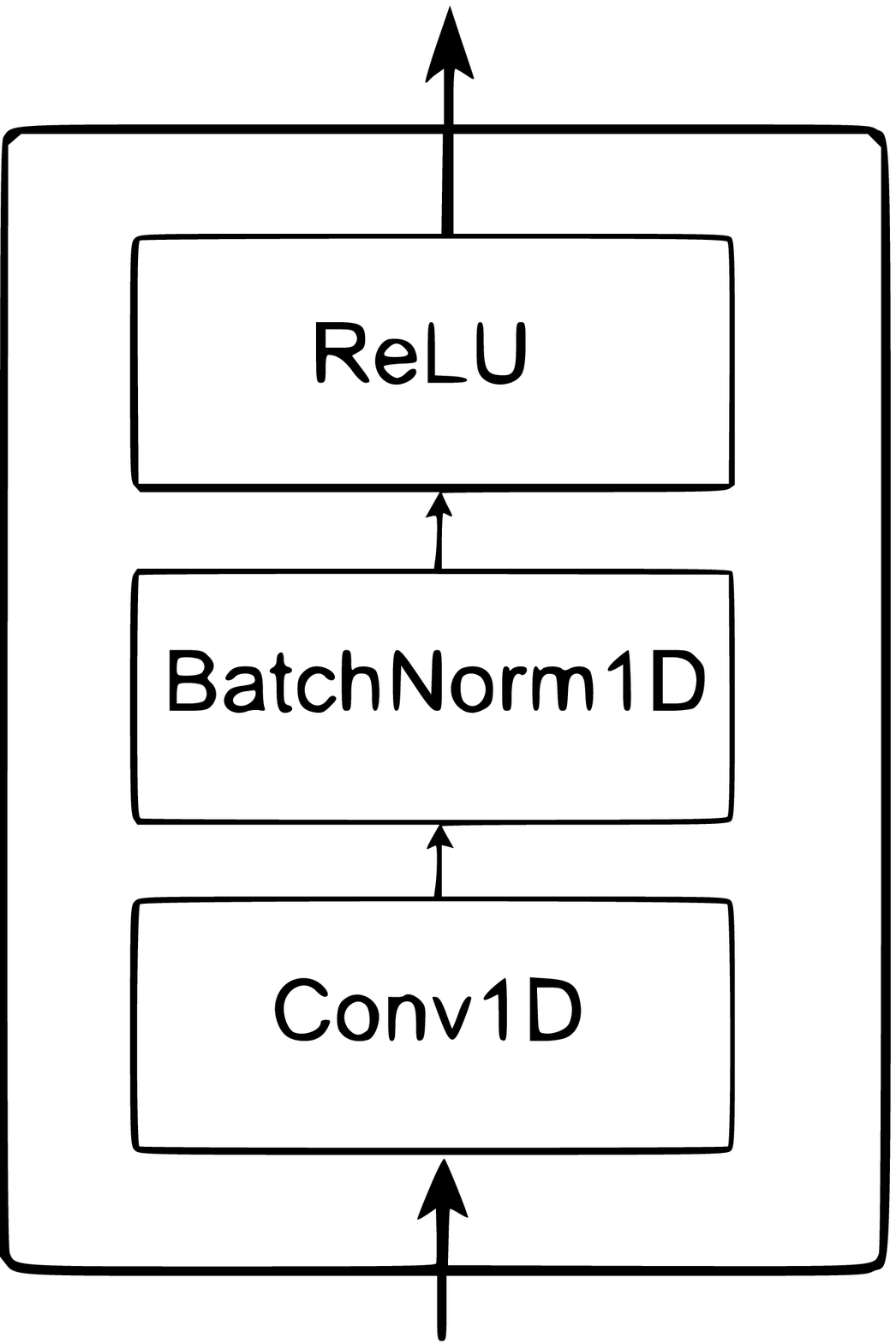}
    \caption{Conv block}
    \label{fig:convblock}
  \end{subfigure}%
  \caption{The architecture of the attack model consisting of three convolution blocks, one-dimensional batch normalization and a final linear layer with binary membership prediction.}
  \label{fig:attack_model_arch}
\end{figure}

\subsection*{Table for Section 3}

\begin{table}[ht]
\centering
\begin{tabular}{|c|c|c|}
\hline
\textbf{The knowledge of $y$}& \textbf{Attack method}   & \textbf{Attack accuracy} \\ \hline
\ding{51} &True label               & 87.9\%                    \\ \hline
\ding{55} &Entropy                  & 78.6\%                    \\ \hline
\ding{55} &Maximum prediction value & 77.4\%                    \\ \hline
\end{tabular}
\caption{The performance of passive membership inference attacks in FL with 4 participants after T=300 epochs for classifying CIFAR100 datset using AlexNet under different assumptions on the adversary capability to access the label $y$ in $\traindataset$. The observed epochs are the last 10 epochs.} 
\label{tab: without_konwledge_y}
\end{table}

\subsection*{Experiment details}
\paragraph{Federated learning setup}
The federated learning setup is set to the same as the open-source code provided for~\cite{nasr}.~\footnote{\url{https://github.com/SPIN-UMass/MembershipWhiteboxAttacks/blob/master/ATTACK-ALEXNET-grad_fed_local.py}.}  
For CIFAR100 dataset, at each communication round, each client performs one local epoch update using SGD optimizer with batch size 100. The learning rate is set to 0.05 for the first 20 epochs, 0.005 for epochs from 21 to 40 and 0.0005 for the epochs from 41 to 300.
For Purchase100 dataset, at each communication round, each client performs one local epoch update using Adam optimizer with batch size 100 and learning rate $0.001$.

\paragraph{Our attack setup} To train the attack model,  we use the Adam optimizer with batch size 100 and learning rate of 0.001. The model is trained for 100 epochs. 
For both CIFAR100 and Purchase100 dataset, the model is trained on 2000 members and 2000 non-members data samples and tested on 5000 members and 5000 non-members data samples. Notice that, compared with  the baseline~\cite{nasr},  we train on less samples but test on the same numbers of data samples.

\end{document}